\documentclass[11pt]{article}

\usepackage[utf8]{inputenc}
\usepackage[T1]{fontenc}
\usepackage{amsmath}
\usepackage{amssymb}
\usepackage{booktabs}
\usepackage{graphicx}
\usepackage{hyperref}
\usepackage{microtype}
\usepackage{enumitem}

\IfFileExists{neurips_2025.sty}{
  \usepackage[preprint]{neurips_2025}
}{
  \usepackage{geometry}
  \geometry{margin=1in}
}

\title{Generic Triple-Latent Compression with Gated Associative Retrieval}
\author{
Liu Xiao \\
\texttt{liu.xiao.in@gmail.com}
}
\date{}

\begin{document}
\maketitle

\begin{abstract}
Can compressed latent state recover higher-order token interactions without
explicitly enumerating token triplets or relying on benchmark-specific parsing?
We study a small family of \textbf{generic triple-latent} sequence models that
maintain a running token state together with a compressed pair-memory pathway.
The resulting base architecture is recurrent, untyped, and trained only from
raw token sequences. We first evaluate three variants---dense pair memory
(\texttt{triple-latent}), slot-compressed pair memory (\texttt{triple-slot}),
and pair memory plus local convolution (\texttt{triple-hybrid})---against a
standard Transformer baseline on byte-level WikiText-2, associative recall, and
throughput scaling. In a matched-width setting, all three latent variants
improve validation bits-per-byte over the Transformer baseline, with
\texttt{triple-hybrid} reaching 4.766 versus 5.124. In an approximately
parameter-matched language-model follow-up, \texttt{triple-hybrid} still
improves on the Transformer, reaching 5.067 at 169{,}424 parameters versus
5.124 at 174{,}848 parameters. A targeted near-budget sweep finds an even
stronger winner: \texttt{triple-hybrid} with $d=64$ and state dimension $8$
reaches 4.827 at 177{,}752 parameters, and averages 4.806 across three seeds
versus 5.143 for the Transformer baseline. The latent-only models do not
improve retrieval-heavy associative recall, so we add a recall-focused hybrid:
separate key-value memory, previous-token queries, last-layer retrieval, and a
gated late-fusion path. This gated hybrid reaches 41.9\% mean associative
recall across three seeds at $d=64$, compared with 25.0\% for the Transformer
and 11.6\% for plain \texttt{triple-hybrid}, and attains 100.0\% in its best
seed. We also test a tokenizer-based MiniMind language-model benchmark under a
shared three-seed, $120$-step protocol. There, the Transformer baseline reaches
mean token loss 7.317, \texttt{transformer\_plus\_triple} reaches 7.239,
\texttt{triple-hybrid} reaches 7.035, and the recall-focused gated hybrid
reaches 6.766. The gain is seed-sensitive and the models remain far slower than
attention in the current Python reference implementation. A later warmup-plus-
curriculum retraining attempt collapses the gated model back to 12.7\% mean
recall. We interpret these results as evidence that latent compression helps
generic language modeling, including on a tokenized pretraining benchmark,
while exact retrieval benefits from a separate gated memory path rather than a
single compressed state alone. Code and experiment scripts are available at
\url{https://github.com/xiaol/Autoresearch_ideas.git}.
\end{abstract}

\section{Introduction}
Higher-order token interactions sit at the heart of sequence modeling, but the
standard way to capture them---full self-attention---comes with quadratic
sequence interactions and a strong dependence on highly optimized kernels
\citep{vaswani2017attention}. Recurrent and linear-time alternatives aim to
replace that cost with compressed state, but often lose expressive power in the
process.

This paper studies one narrow question: can a generic latent state compress
\emph{triple-like} token interactions well enough to help on a standard
language-model benchmark? This question is motivated by two observations. First,
matrix-state and hybrid recurrent models suggest that richer state updates can
recover more of the behavior usually attributed to attention
\citep{m2rnn2026, eaglefinch2024}. Second, mechanism-specific synthetic
benchmarks, while useful, are not sufficient evidence that an architecture
helps on generic sequence modeling.

We therefore focus on a deliberately simple benchmark stack: byte-level
WikiText-2 language modeling \citep{merity2017pointer}, a tokenizer-based
MiniMind pretraining benchmark, a synthetic associative-recall stress test, and
sequence-length throughput profiling on Apple MPS. Our model family is
intentionally generic. It receives only token IDs, does not parse typed roles,
and uses no task-specific symbolic heads.

The paper now reports three linked stories. First, the base triple-latent
family already improves generic byte-level language modeling over a small
Transformer. Second, these gains transfer to a tokenized MiniMind benchmark,
including a \texttt{transformer\_plus\_triple} compatibility model that keeps
standard self-attention lookup and adds triple interaction only as a late
readout branch. Third, the original latent-only models fail on associative
recall, and the best fix is not to force exact retrieval into the same
compressed state. Instead, a recall-focused hybrid adds a separate key-value
path with gated late fusion.

\paragraph{Contributions.}
\begin{itemize}[leftmargin=1.25em]
  \item We define a generic triple-latent architecture family that combines a
  running token state with a compressed pair-memory pathway.
  \item We extend this family with a recall-focused hybrid that keeps the
  triple-latent path for compression while adding separate key-value memory,
  previous-token queries, and gated late fusion for retrieval.
  \item We provide a dedicated architecture figure and a standalone benchmark
  driver in the repository.
  \item We show that all three matched-width triple-latent variants improve
  small-scale byte-level WikiText-2 over a Transformer baseline, and that the
  strongest \texttt{triple-hybrid} variant still improves after approximate
  parameter matching.
  \item We add a tokenizer-based MiniMind benchmark and show that all tested
  triple-augmented models improve over the Transformer on mean token loss;
  \texttt{transformer\_plus\_triple} provides a cleaner compatibility path,
  while the gated hybrid is the strongest LM-quality model in this small
  benchmark.
  \item We show that the latent-only family fails on associative recall, but a
  gated hybrid retrieval path can outperform the Transformer on this benchmark
  on average across three seeds, albeit with substantial seed sensitivity.
  \item We show the corresponding systems caveat clearly: none of these models
  achieve a wall-clock throughput win in the current reference implementation.
\end{itemize}

\paragraph{Code availability.}
The research code, experiment drivers, and paper sources are available at
\url{https://github.com/xiaol/Autoresearch_ideas.git}.

\section{Related Work}
The first backdrop is the growing literature on efficient alternatives to
quadratic attention. Depth-recurrent architectures such as Universal
Transformers \citep{dehghani2019universal}, linear-attention formulations that
recover recurrent updates \citep{katharopoulos2020transformers}, retention and
state-space replacements such as RetNet and Mamba
\citep{retnet2023, mamba2023}, and recent recurrent revivals such as
RWKV, Gated DeltaNet, xLSTM, and matrix-state language models
\citep{rwkv2023, gateddeltanet2026, beck2024xlstm, m2rnn2026, eaglefinch2024}
all study how much of
attention's behavior can be recovered with compressed state and linear-time
sequence interaction. Our base triple-latent family belongs to this line, but
explicitly separates a running token state from a compressed pair-memory path
targeted at higher-order interactions.

The second backdrop is explicit memory and retrieval augmentation. Product-key
memory, nearest-neighbor language models, and Memorizing Transformers show that
exact or approximate associative lookup can complement parametric sequence
modeling \citep{lample2019productkeys, khandelwal2019knnlm, memorizing2022}.
Our recall-focused hybrid is closest in spirit to this family and to recent
hybrid associative memory formulations \citep{ham2026}, but it is deliberately
minimal: retrieval is added only at the output layer, and the lossy recurrent
compression path is left untouched.

Finally, recent production-oriented hybrids often combine multiple sequence
mechanisms rather than committing to a single primitive. Compressive
Transformers, Jamba, Samba, and Hymba illustrate different ways to mix local
compression, attention, and recurrent or state-space components
\citep{compressive2019, jamba2024, samba2024, hymba2024}. Our goal here is
smaller-scale and more diagnostic: use a controlled benchmark stack to separate
what latent compression helps from what exact retrieval helps in a single
generic architecture.

\section{Method}
\subsection{Generic Triple-Latent Layer}
\begin{figure}[t]
  \centering
  \includegraphics[width=\linewidth]{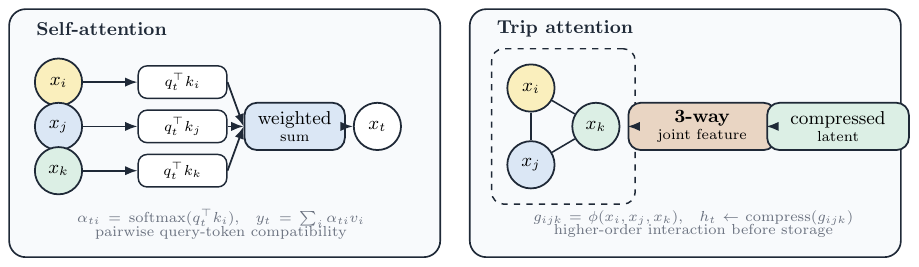}
  \caption{Self-attention computes pairwise query-token compatibilities and
  aggregates values with a weighted sum. Trip attention instead forms an
  explicit higher-order feature from a token triplet before that interaction is
  compressed into a latent state.}
  \label{fig:trip-vs-self-attention}
\end{figure}

Each layer takes token features $x_t \in \mathbb{R}^d$ and produces four sets
of projections:
\begin{equation}
a_t, b_t, q_t^\ell, q_t^r \in \mathbb{R}^{H \times S},
\end{equation}
along with learned decay gates for a running state $s_t$ and a compressed
pair-memory pathway. The state update is
\begin{equation}
s_t = \lambda_t \odot s_{t-1} + (1 - \lambda_t) \odot a_t,
\end{equation}
where $\lambda_t$ is a learned per-token retention gate. Dense pair memory then
stores an interaction between the \emph{previous} state and the current write
vector:
\begin{equation}
P_t = \gamma_t \odot P_{t-1} + (1 - \gamma_t) \odot (s_{t-1} \otimes b_t),
\end{equation}
with $P_t \in \mathbb{R}^{H \times S \times S}$. Readout first probes the
compressed pair memory with a left query and then composes the result with a
right query:
\begin{equation}
o_t = W_o \, \mathrm{vec}\!\left((P_t q_t^\ell) \odot q_t^r\right).
\end{equation}
The layer output adds this mixed readout back to the residual stream and then
applies an FFN block.
Figure~\ref{fig:trip-vs-self-attention} contrasts this higher-order pathway
with ordinary self-attention: self-attention scores one query against one key at
a time, while the triple-latent path explicitly forms a joint feature from a
token triplet before compression.

\subsection{Variants}
We study three generic variants.
\begin{itemize}[leftmargin=1.25em]
  \item \textbf{Triple-Latent}: dense pair memory $P_t$.
  \item \textbf{Triple-Slot}: replace dense pair memory with learned slots that
  store low-rank left and right factors.
  \item \textbf{Triple-Hybrid}: keep dense pair memory and add a local
  convolutional path before the recurrent stack.
\end{itemize}

Unlike benchmark-aware typed latent models, these variants do not expose
special roles, tags, or symbolic task heads.

\subsection{Transformer+Triple Compatibility Baseline}
To separate ``triple interaction helps'' from ``replace attention entirely,''
we also evaluate a compatibility baseline that keeps a standard Transformer
stack and adds one triple-latent layer only at the output. Let $h_t$ be the
final Transformer hidden state and let $e_t$ denote the token-embedding stream.
We run a single triple-latent layer with token-sourced left keys and no
associative or direct-lookup branch to produce a higher-order readout $c_t$.
The final logits are
\begin{equation}
\ell_t = W_{\mathrm{vocab}} h_t + W_{\mathrm{vocab}}\!\left(\mathrm{LN}(c_t)\right).
\end{equation}
This \texttt{transformer\_plus\_triple} model preserves ordinary
self-attention key-value lookup and treats the triple pathway as an additive
high-order residual head. We use it only as a diagnostic compatibility
baseline in the MiniMind and associative-recall follow-ups.

\subsection{Gated Hybrid Associative Retrieval}
The base triple-latent family compresses context well, but it does not provide
explicit content-addressable lookup. To test whether exact retrieval should be
handled separately, we add a recall-focused hybrid that keeps the triple-latent
stack and augments only the top layer with an explicit key-value memory.

\begin{figure}[t]
  \centering
  \includegraphics[width=\linewidth]{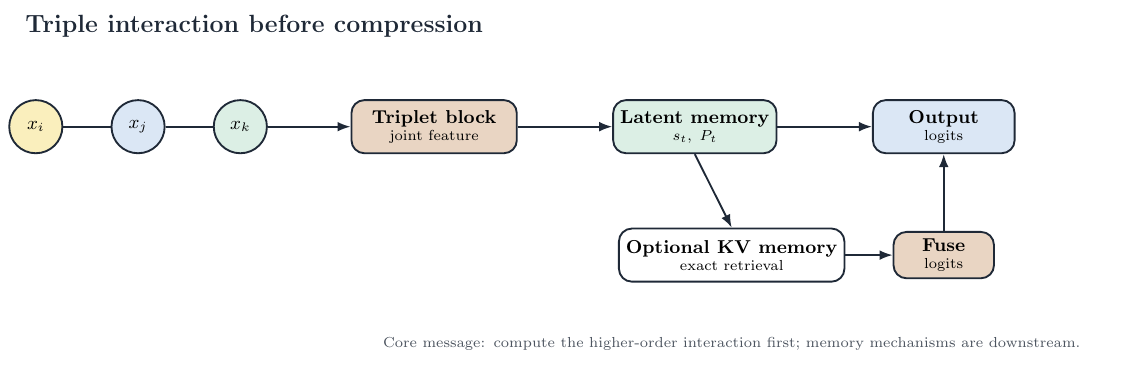}
  \caption{Simple intuition for the generic triple-latent family. Three tokens
  interact jointly through an explicit higher-order mixing step rather than
  only through pairwise attention scores. The resulting interaction is then
  written into a compressed recurrent latent path, while an optional exact
  key-value branch can be kept separate and fused only at the output.}
  \label{fig:architecture}
\end{figure}

Given token features $e_t$, the retrieval path writes key-value pairs
$(k_{t-1}, v_t)$ into an append-only memory using learned projections
\begin{equation}
k_t = W_k e_t, \qquad v_t = W_v e_t.
\end{equation}
At retrieval time, we query the memory using the \emph{previous-token}
representation,
\begin{equation}
q_t = W_q e_{t-1},
\end{equation}
and compute top-$k$ similarity scores against stored normalized keys:
\begin{equation}
\alpha_{t,i} \propto \exp\!\left(\frac{\hat{q}_t^\top \hat{k}_i}{\sqrt{d_a}}\right).
\end{equation}
The retrieved memory vector is
\begin{equation}
m_t = \sum_i \alpha_{t,i} v_i.
\end{equation}
We then apply a learned read gate and fuse only at the output layer:
\begin{equation}
\tilde{m}_t = \sigma(W_g q_t) \odot m_t,
\qquad
\ell_t = W_{\mathrm{vocab}} h_t + W_{\mathrm{vocab}}\!\left(\mathrm{LN}(\tilde{m}_t)\right).
\end{equation}
This design leaves the triple-latent state untouched by retrieval writes and
keeps exact lookup separate from lossy recurrent compression.
Figure~\ref{fig:architecture} summarizes this separation: triple interaction is
computed first, then stored through a compressed latent path, while exact
retrieval remains a downstream optional branch.

\section{Experimental Setup}
\paragraph{Language modeling.}
We train byte-level autoregressive models on WikiText-2 raw text
\citep{merity2017pointer}. Inputs are UTF-8 bytes with a 259-token vocabulary.
The matched-width setting uses $d=64$, $3$ layers, $4$ heads, sequence length
$128$, batch size $16$, and $80$ optimizer steps on Apple MPS.

\paragraph{MiniMind tokenizer-based LM.}
To test whether the byte-level gain transfers to a more standard tokenized
setting, we also benchmark on the MiniMind pretraining corpus and tokenizer
shipped in the repository. We follow the same small standard protocol used by
the existing MiniMind harness: hidden size $96$, $2$ layers, $4$ heads,
sequence length $96$, batch size $4$, $120$ optimizer steps, $32$ evaluation
batches, learning rate $3 \times 10^{-4}$, deterministic $1$-in-$20$ holdout,
and seeds $\{11,22,33\}$. The compared models are Transformer,
\texttt{triple-latent}, \texttt{triple-hybrid},
\texttt{transformer\_plus\_triple}, and the recall-focused gated hybrid.

\paragraph{Associative recall.}
We also train each model on a synthetic key-value retrieval task with four
key-value pairs, random filler, and a final query. Models run for $200$ steps
with batch size $32$ and sequence length $128$. Accuracy is measured on the
final answer token only.

\paragraph{Recall-focused hybrid follow-up.}
Because the original latent-only family and early no-gate dual-memory variants
remained near chance on associative recall, we run a second memory study for
the gated hybrid described above. We sweep widths $d \in \{64,80,96\}$ at
$500$ associative-recall training steps, then run three seeds
($7,19,29$) for the best width against same-width Transformer and plain
\texttt{triple-hybrid} baselines. We also report byte-level WikiText-2 and
throughput for the best recall-focused configuration under the same
$80$-step LM schedule used elsewhere in the paper.

\paragraph{Stabilization follow-up.}
We also test whether the gated winner can be made more reliable with simple
optimization changes alone. This follow-up keeps the same architecture and
uses a $50$-step learning-rate warmup, a sequence-length curriculum from $32$
to $128$ over the first $150$ steps, and read-gate bias initialization at
$-2.0$. We again report a three-seed mean at $d=64$.

\paragraph{Parameter-matched follow-up.}
To check whether the WikiText-2 gain is only a width effect, we run an
approximate parameter-matched language-model follow-up using the same optimizer
schedule but reduced widths: \texttt{triple-latent} and \texttt{triple-slot}
use $d=60$, and \texttt{triple-hybrid} uses $d=56$. These settings place all
three variants near the Transformer's 174{,}848 parameters.

\paragraph{Throughput benchmark.}
We measure forward-only inference on Apple MPS at sequence lengths
$\{64,128,256,512\}$ with batch size $8$ and $20$ timed iterations. The
Transformer benefits from optimized attention kernels, while the triple-latent
models remain Python-loop reference implementations.

\paragraph{Compute accounting.}
The main training comparisons in the paper are still step-matched rather than
FLOP-matched. To quantify the arithmetic profile more clearly, we also estimate
dominant forward FLOPs analytically from the implemented model definitions.
The estimate counts the main matrix multiplies and tensor contractions, the
\texttt{triple-hybrid} local convolution, and any extra vocabulary projection
from late logit fusion. It omits embedding lookups, normalization, softmax,
top-$k$ selection, and Python overhead.

\section{Results}
\subsection{Generic Benchmark}
Table~\ref{tab:generic} is the main result for the \emph{base} triple-latent
family. In the matched-width setting, all three triple-latent variants beat the
Transformer on byte-level WikiText-2. The strongest \texttt{triple-hybrid}
variant reaches 4.766 bits-per-byte versus 5.124 for the Transformer baseline.
This improvement is not entirely explained by parameter count: in the
approximate parameter-matched follow-up, \texttt{triple-hybrid} still reaches
5.067 bits-per-byte at fewer parameters than the Transformer.

We then pushed the search further under a near-baseline parameter budget.
A targeted sweep over \texttt{triple-hybrid} and \texttt{triple-slot-hybrid}
variants found the strongest fair-budget configuration at
\texttt{triple-hybrid} with $d=64$ and state dimension $8$. That model uses
177{,}752 parameters, reaches 4.827 bits-per-byte in the pilot run, and
achieves 4.806 mean bits-per-byte across three seeds. The corresponding
Transformer baseline averages 5.143 across the same three seeds. This makes the
generic result stronger than the earlier single-run comparison alone: the
quality win survives both a targeted budget sweep and a small robustness check.

\begin{table}[t]
\centering
\small
\begin{tabular}{llcccc}
\toprule
Regime & Model & Params & Val. BPB $\downarrow$ & Val. PPL $\downarrow$ & Recall $\uparrow$ \\
\midrule
Matched width & Transformer & 174{,}848 & 5.124 & 34.88 & \textbf{25.4\%} \\
Matched width & Triple-Latent & 187{,}928 & 5.022 & 32.50 & 13.4\% \\
Matched width & Triple-Slot & 188{,}024 & 5.041 & 32.92 & 13.4\% \\
Matched width & Triple-Hybrid & 208{,}472 & \textbf{4.766} & \textbf{27.21} & 13.4\% \\
Parameter-matched LM & Triple-Latent & 170{,}424 & 5.114 & 34.64 & -- \\
Parameter-matched LM & Triple-Slot & 170{,}520 & 5.126 & 34.92 & -- \\
Parameter-matched LM & Triple-Hybrid & \textbf{169{,}424} & \textbf{5.067} & \textbf{33.52} & -- \\
Budget winner & Triple-Hybrid ($64,8$) & 177{,}752 & \textbf{4.806} & \textbf{27.98} & 13.4\% \\
\bottomrule
\end{tabular}
\caption{Generic triple-latent results. All three matched-width variants
improve over the Transformer baseline on byte-level WikiText-2. The strongest
\texttt{triple-hybrid} variant still improves after approximate parameter
matching, and a targeted near-budget sweep yields an even stronger three-seed
winner. None of the \emph{latent-only} variants improve associative recall.}
\label{tab:generic}
\end{table}

\subsection{MiniMind Tokenizer-Based Benchmark}
The byte-level WikiText-2 result still leaves an external-validity question:
does the gain survive on a more standard tokenized language-model benchmark?
Table~\ref{tab:minimind} says yes. On the MiniMind benchmark, all tested
triple-augmented models beat the plain Transformer on mean token loss. The
strongest is the gated hybrid at 6.766 versus 7.317 for the Transformer, while
plain \texttt{triple-hybrid} remains strong at 7.035.

The compatibility baseline is especially informative here.
\texttt{transformer\_plus\_triple} reaches 7.239 mean token loss and 5.21\%
token accuracy, compared with 7.317 and 4.28\% for the plain Transformer. So
the triple interaction signal still helps even when the standard attention
key-value path is kept intact. At the same time, it is not the strongest LM
model in this family: the more aggressive \texttt{triple-hybrid} variants gain
more on quality, while paying a larger systems cost.

\begin{table}[t]
\centering
\small
\begin{tabular}{lcccc}
\toprule
Model & Params (M) & Token loss $\downarrow$ & Token acc $\uparrow$ & Train time rel. $\downarrow$ \\
\midrule
Transformer & 0.920 & 7.317 & 4.28\% & 1.0$\times$ \\
Triple-Latent & 0.867 & 7.295 & 5.28\% & 10.8$\times$ \\
Transformer+Triple & 1.051 & 7.239 & 5.21\% & 5.8$\times$ \\
Triple-Hybrid & 0.913 & 7.035 & 6.07\% & 11.0$\times$ \\
Gated Hybrid Retrieval & \textbf{0.956} & \textbf{6.766} & \textbf{7.17\%} & 35.9$\times$ \\
\bottomrule
\end{tabular}
\caption{MiniMind tokenizer-based language-model benchmark. Results are
three-seed means under a shared $120$-step protocol (hidden size $96$, $2$
layers, sequence length $96$, batch size $4$, deterministic $1$-in-$20$
holdout). All tested triple-augmented models outperform the plain Transformer
on mean token loss. \texttt{transformer\_plus\_triple} offers a cleaner
compatibility path that preserves self-attention lookup, while the gated hybrid
is the strongest LM-quality model in this small benchmark.}
\label{tab:minimind}
\end{table}

\subsection{Retrieval Ablation}
Before adding gating, we tested whether the recall failure was simply a query,
fusion, or layer-placement problem. Table~\ref{tab:retrieval_ablation} shows
that the answer is no. Parallel, serial, and previous-token queries all remain
at 13.4\% accuracy in the short dual-memory sweeps. Last-layer-only retrieval
helps efficiency relative to all-layer retrieval, and logit fusion helps byte-
level language modeling much more than sum or concat fusion, but these changes
alone do not move associative recall.

The critical transition is therefore not ``use previous-token query'' by
itself. It is ``use previous-token query, read only at the last layer, fuse at
the logits, and add a learned read gate.'' A same-budget long-run check of an
ungated previous-query model also stays near the old failure mode at 11.96\%
accuracy, which reinforces this conclusion.

\begin{table}[t]
\centering
\small
\begin{tabular}{lcccccc}
\toprule
Variant & Query & Layers & Fusion & Gate & Recall $\uparrow$ & Val. BPB $\downarrow$ \\
\midrule
Transformer & -- & -- & -- & -- & 25.4\% & 5.144 \\
Triple-Hybrid & -- & -- & -- & -- & 13.4\% & 4.843 \\
Assoc sum & par. & all & sum & no & 13.4\% & 4.835 \\
Assoc concat & par. & all & concat & no & 13.4\% & 4.767 \\
Assoc logits & par. & all & logits & no & 13.4\% & 3.664 \\
Serial sum & serial & all & sum & no & 13.4\% & 4.834 \\
Prev. logits & prev. & all & logits & no & 13.4\% & 3.657 \\
Last logits & par. & last & logits & no & 13.4\% & 3.680 \\
Gated winner & prev. & last & logits & yes & \textbf{41.9\%} & \textbf{3.595} \\
\bottomrule
\end{tabular}
\caption{Retrieval ablation for the recall-focused hybrid family at $d=64$.
Ungated rows come from the $200$-step dual-memory sweeps, while the final row
reports the $500$-step three-seed mean of the gated winner. Query and fusion
choices affect LM compression much more than recall; the first configuration
that changes recall materially is previous-token query plus last-layer
retrieval plus a read gate and late logit fusion.}
\label{tab:retrieval_ablation}
\end{table}

\subsection{Recall-Focused Hybrid}
The ablation above narrows the search space but still leaves a clear gap:
compression improves, but exact retrieval does not. We therefore ran a second,
recall-focused follow-up on the previous-query, last-layer, logit-fusion path
with an added read gate. This is the first configuration that changes the
benchmark meaningfully.

Table~\ref{tab:recall_hybrid} summarizes the best recall-focused model at
$d=64$. Across three seeds, \texttt{triple-hybrid-assoc-gated-last-logits}
reaches 41.9\% mean associative-recall accuracy, compared with 25.0\% for the
Transformer and 11.6\% for plain \texttt{triple-hybrid}. The best seed solves
the task at 100.0\% accuracy. Just as importantly, longer training alone does
not explain the gain: same-budget ungated baselines remain near 12\%.

The result is encouraging but not yet fully stable. The recall gain is highly
seed-sensitive: one seed converges to exact retrieval, while two remain near
the old failure mode. Width scaling is also non-monotonic in the current
implementation: a single-seed sweep gives 100.0\% at $d=64$, 67.2\% at
$d=80$, and a collapse back to 11.96\% at $d=96$. We therefore treat the
gated hybrid as a promising direction rather than a solved replacement.

A later optimization-only stabilization attempt strengthens that caution rather
than weakening it. The warmup-plus-curriculum retraining described above
reaches only 12.7\% mean recall across the same three seeds, with a best seed
of 13.4\%. In other words, the original high-recall mode is not reproduced by
simple schedule shaping or gate-bias initialization alone.

A compatibility baseline that preserves standard attention lookup also helps
clarify the recall story. In a short $120$-step recall screen at $d=64$,
\texttt{transformer\_plus\_triple} reaches 21.9\% accuracy versus 19.1\% for
the plain Transformer, which suggests that additive triple features can help
early retrieval learning without disrupting the key-value path. In the longer
$500$-step follow-up, however, the Transformer recovers to 24.6\% and slightly
surpasses \texttt{transformer\_plus\_triple} at 23.4\%. We therefore treat
\texttt{transformer\_plus\_triple} as a promising compatibility baseline, not
yet the final recall winner.

\begin{table}[t]
\centering
\small
\begin{tabular}{lccccc}
\toprule
Model & Recall mean $\uparrow$ & Best seed $\uparrow$ & Val. BPB $\downarrow$ & Tok/s 64 $\uparrow$ & Tok/s 512 $\uparrow$ \\
\midrule
Transformer & 25.0\% & 25.6\% & 5.144 & 602{,}684 & 344{,}185 \\
Triple-Hybrid & 11.6\% & 12.0\% & 4.843 & 18{,}433 & 18{,}870 \\
Gated Hybrid Retrieval & \textbf{41.9\%} & \textbf{100.0\%} & \textbf{3.595} & 5{,}659 & 4{,}910 \\
\bottomrule
\end{tabular}
\caption{Recall-focused gated hybrid at $d=64$. Recall numbers are three-seed
means over $500$ associative-recall steps, with the best single-seed run shown
separately. Byte-level WikiText-2 and throughput use the same $80$-step LM
setup as the rest of the paper. The gated hybrid improves both recall and LM
compression, but remains slower than the plain triple-hybrid and much slower
than the Transformer.}
\label{tab:recall_hybrid}
\end{table}

\subsection{Approximate Compute Accounting}
The training runs in this paper are still step-matched, not FLOP-matched, but
an analytic forward estimate is still informative. Table~\ref{tab:compute}
counts the dominant tensor contractions from the implemented models. At the
training context of $128$, plain \texttt{triple-hybrid} uses about $0.80\times$
the Transformer's dominant forward FLOPs, and the gated hybrid uses about
$0.95\times$. At sequence length $512$, these estimates fall to $0.47\times$
and $0.60\times$, respectively.

The wall-clock result is the opposite because arithmetic count is not the whole
story in the current codebase. The Transformer runs through optimized attention
kernels, while the triple-latent models execute recurrent Python loops and
unfused retrieval logic. The gated model also reintroduces quadratic
sequence-length cost through associative scoring, even though its constant is
smaller than full attention in this small configuration.

\begin{table}[t]
\centering
\small
\begin{tabular}{lccc}
\toprule
Model & Seq 64 & Seq 128 & Seq 512 \\
\midrule
Transformer & 24.1 (1.00x) & 54.6 (1.00x) & 369.3 (1.00x) \\
Triple-Hybrid & 21.8 (0.90x) & 43.5 (0.80x) & 174.0 (0.47x) \\
Gated Hybrid Retrieval & 25.8 (1.07x) & 52.1 (0.95x) & 220.9 (0.60x) \\
\bottomrule
\end{tabular}
\caption{Approximate forward MFLOPs from the dominant tensor contractions in
the repo implementation. Counts include the hybrid local convolution and the
gated model's extra logits head, but omit embedding lookups, normalization,
softmax, top-$k$ selection, and Python overhead.}
\label{tab:compute}
\end{table}

\subsection{Memory and Throughput Tradeoffs}
The expanded results now come with two clear caveats.

First, explicit retrieval is no longer a simple failure story, but it is still
not cleanly solved. The latent-only variants remain at 13.4\% after the
original short runs, and the best gated hybrid is strongly seed-sensitive even
though its three-seed mean surpasses the Transformer. A later warmup-plus-
curriculum retraining attempt falls back to 12.7\% mean recall, so there is
still no simple recipe that stabilizes the high-recall mode.

Second, the current implementation is still not a systems win. The plain
\texttt{triple-hybrid} is about $32.7\times$ slower than the Transformer at
sequence length $64$ and $18.2\times$ slower at length $512$. The recall-
focused gated hybrid is about $106.5\times$ and $70.1\times$ slower at the
same lengths, despite the more favorable arithmetic profile in
Table~\ref{tab:compute}. The recurrent path plus explicit retrieval are both
still unfused Python reference code. The same pattern appears on the MiniMind
benchmark: \texttt{transformer\_plus\_triple}, \texttt{triple-hybrid}, and the
gated hybrid take about $5.8\times$, $11.0\times$, and $35.9\times$ the
Transformers training time, respectively.

\section{Discussion}
The central lesson is that generic latent compression appears to be real, but
exact retrieval should not be forced into the same compressed state. The
positive language-model result is not limited to a benchmark-aware symbolic
construction: a fully untyped latent family improves a standard byte-level
language-model benchmark, and the strongest variant survives approximate
parameter matching. The same direction also transfers to the tokenized
MiniMind benchmark, where every tested triple-augmented model improves over the
Transformer on mean token loss. That makes the compression story more credible
than a synthetic benchmark win alone.

The \texttt{transformer\_plus\_triple} baseline sharpens that interpretation.
It keeps a clean self-attention lookup path, improves both early recall and
MiniMind token loss over the Transformer, and therefore suggests that triple
interaction is useful even when it is not asked to replace attention outright.
But it is not yet the strongest model either on recall or on MiniMind quality,
so at present it looks more like a compatibility path than a final endpoint.

The memory result sharpens the picture further. A separate key-value path with
late gated fusion can recover associative recall and even outperform the
Transformer on average in our three-seed $d=64$ follow-up. But the behavior is
still unstable across seeds and widths. A later warmup-plus-curriculum
retraining attempt also fails to reproduce the original high-recall mode, which
suggests that the remaining problem is not merely a missing warmup schedule.
This suggests that the next research steps should focus on two fronts:
\begin{itemize}[leftmargin=1.25em]
  \item better optimization and routing for the gated retrieval path, so the
  high-recall solution becomes a stable mode rather than an occasional one,
  and so it transfers beyond the dedicated associative-recall benchmark
  \item fused scan, fused retrieval, and custom recurrent kernels, so the
  architecture can realize its favorable arithmetic profile in wall-clock terms
\end{itemize}

\section{Limitations}
This is a pilot study with several important limits.
\begin{itemize}[leftmargin=1.25em]
  \item Most experiments are small Apple MPS runs. The key recall result and
  the MiniMind follow-up both use three seeds, but the broader sweep is still
  far from exhaustive.
  \item We compare matched training steps, not matched training FLOP budgets.
  Table~\ref{tab:compute} is an analytic forward estimate, not a full training
  compute accounting.
  \item The MiniMind benchmark is more realistic than the byte-level stack, but
  it is still a short-horizon pretraining proxy: one corpus, one tokenizer, and
  only $120$ update steps.
  \item An exploratory zero-shot RULER-core check remained at 0.0 at context
  lengths $128$ and $256$ for the small byte-level checkpoints used here. We
  therefore do not treat RULER as a main result in this paper, and we do not
  yet report broader long-context suites such as LongBench, full RULER,
  BABILong, and NoLiMa
  \citep{longbench2023, hsieh2024ruler, babilong2024, nolima2025}.
  \item The gated associative-recall improvement is seed-sensitive, with one
  seed solving the task and others remaining much weaker. Our optimization-only
  stabilization follow-up does not recover that strong mode.
  \item The throughput numbers come from a Python-loop reference
  implementation, not an optimized recurrent kernel.
\end{itemize}

\section{Conclusion}
Generic triple-latent compression is strong enough to merit its own line of
investigation, but the updated story is now two-part. First, the base
triple-latent family improves generic byte-level language modeling over a small
Transformer baseline, and the best \texttt{triple-hybrid} remains competitive
under approximate parameter matching. These gains also transfer to a tokenized
MiniMind benchmark: all tested triple-augmented models beat the Transformer on
mean token loss, with the gated hybrid strongest and
\texttt{transformer\_plus\_triple} providing a cleaner compatibility path.
Second, exact retrieval becomes substantially stronger when we add a separate
gated key-value path on top of the triple-latent stack: the best recall-
focused hybrid beats the Transformer on mean associative recall in our
three-seed $d=64$ follow-up and reaches 100.0\% in its best run. The result is
still unstable across seeds and much slower than attention. An optimization-
only stabilization attempt also fails to reproduce the original high-recall
mode. So this is not yet a practical replacement for standard attention. But it
is meaningful evidence that latent compression and exact retrieval can coexist
in a single generic architecture when they are routed through separate
pathways.

\appendix
\section{Implementation Notes for a Competitive Kernel}
The current codebase is a research reference implementation, not a production
kernel stack. A serious systems comparison against modern attention would need
at least four engineering upgrades.
\begin{itemize}[leftmargin=1.25em]
  \item \textbf{Fused recurrent scan.} The state update, pair-memory update,
  and readout should run as a single fused scan across sequence positions
  rather than a Python loop over tokens.
  \item \textbf{On-device associative lookup.} The explicit key-value path
  needs a tiled similarity kernel, top-$k$ selection, and value aggregation on
  device. The current list-append plus stack formulation is simple but
  bandwidth-inefficient.
  \item \textbf{Late-logit fusion without duplicate overhead.} The gated model
  currently pays for an additional vocabulary projection. A stronger systems
  implementation would either fuse the two projections or share more of the
  final readout path.
  \item \textbf{Decode-time cache layout.} Autoregressive decoding needs a
  persistent cache for the recurrent state, pair-memory tensor, and the
  explicit key-value memory. These should be laid out separately so recurrent
  updates stay contiguous while retrieval remains append-only.
  \item \textbf{Compiled backward path.} A training-speed comparison also needs
  fused backward kernels or at least aggressive graph compilation. Without
  this, the measured wall-clock gap says more about implementation maturity
  than about the architecture's arithmetic shape.
\end{itemize}
These requirements explain why the current throughput numbers are much worse
than the FLOP estimates alone would suggest: the baseline Transformer already
benefits from optimized attention kernels, while the triple-latent models do
not yet benefit from equivalent recurrent or retrieval kernels.

\bibliographystyle{plainnat}
\bibliography{refs}

\end{document}